\documentclass[conference]{IEEEtran}
\IEEEoverridecommandlockouts

\usepackage{cite}
\usepackage{amsmath,amssymb,amsfonts}
\usepackage{algorithmic}
\usepackage{graphicx}
\usepackage{textcomp}
\usepackage{xcolor}
\def\BibTeX{{\rm B\kern-.05em{\sc i\kern-.025em b}\kern-.08em
    T\kern-.1667em\lower.7ex\hbox{E}\kern-.125emX}}

\usepackage{hyperref}
\usepackage{booktabs}

\begin{document}

\title{
	Cloud-Aware SAR Fusion for Enhanced Optical Sensing in Space Missions%
	\thanks{This research is partly supported by the National Science and Technology Council (NSTC), Taiwan, under Grant No. 113-2222-E-027-011 and 114-2119-M-011-001.}%
	\thanks{The public source code of the proposed model is available at \url{https://github.com/thoailt/Cloud-Removal}.}
}

\author{
	Trong-An Bui\IEEEauthorrefmark{1}\IEEEauthorrefmark{3}, Thanh-Thoai Le\IEEEauthorrefmark{2} \\
	\IEEEauthorblockA{\IEEEauthorrefmark{1}Institute of Aerospace and System Engineering, National Taipei University of Technology, Taipei, Taiwan}
	\IEEEauthorblockA{\IEEEauthorrefmark{2}Department of Information Technology, Ho Chi Minh City University of Education, Ho Chi Minh City, Vietnam}
	\thanks{\IEEEauthorrefmark{3}Corresponding author: Trong-An Bui (\href{mailto:trongan93@ntut.edu.tw}{trongan93@ntut.edu.tw})}
}

\maketitle
\thispagestyle{empty}
\pagestyle{empty}
\begin{abstract}
	Cloud contamination significantly impairs the usability of optical satellite imagery, affecting critical applications such as environmental monitoring, disaster response, and land-use analysis. This research presents a Cloud-Attentive Reconstruction Framework that integrates SAR-optical feature fusion with deep learning-based image reconstruction to generate cloud-free optical imagery. The proposed framework employs an attention-driven feature fusion mechanism to align complementary structural information from Synthetic Aperture Radar (SAR) with spectral characteristics from optical data. Furthermore, a cloud-aware model update strategy introduces adaptive loss weighting to prioritize cloud-occluded regions, enhancing reconstruction accuracy. Experimental results demonstrate that the proposed method outperforms existing approaches, achieving a PSNR of 31.01\,dB, SSIM of 0.918, and MAE of 0.017. These outcomes highlight the framework’s effectiveness in producing high-fidelity, spatially and spectrally consistent cloud-free optical images.
\end{abstract}

\section{Introduction}
\label{sec:intro}

Cloud cover remains a persistent challenge in optical satellite imaging, especially in tropical and high-humidity regions where frequent and dense clouds obstruct surface visibility. This limitation is particularly critical during natural disasters such as floods and landslides \cite{Andrew2023, Robinson2019}, where rapid and accurate image acquisition is essential for effective response and recovery. Beyond emergency scenarios, cloud-induced data gaps hinder long-term environmental monitoring, land-use planning, and agricultural assessments.

Traditional cloud mitigation strategies rely on two sequential steps: \textit{cloud masking} and \textit{cloud reconstruction}. Cloud masks detect and flag cloud-contaminated pixels based on spectral, thermal, or statistical indicators. Reconstruction methods then estimate missing pixel values using either temporal or spatial information. While temporal interpolation approaches such as STARFM \cite{Gao2006} leverage multi-temporal optical imagery, they falter in persistently cloudy regions where cloud-free observations are unavailable \cite{Hilker2009}. Spatial interpolation methods, like inverse distance weighting and kriging \cite{Roy2010}, are effective in homogeneous areas but struggle in heterogeneous or urban environments \cite{Angel2019}.

An increasingly promising direction is multi-modal data fusion, particularly integrating \textit{Synthetic Aperture Radar (SAR)} and optical imagery. Since SAR operates independently of weather and lighting, it provides structural cues that complement the spectral richness of optical data. Deep learning models have enhanced SAR-optical fusion by learning non-linear relationships across modalities. However, existing approaches often treat all pixels equally during training, overlooking the distinct challenge posed by cloud-occluded regions.

To address this, this research propose a deep learning framework that explicitly prioritizes cloud-covered regions using a \textit{cloud-aware weighting strategy} in the loss function. Our model fuses spatial and spectral features from co-registered SAR and optical inputs through an attention-based mechanism, enabling accurate and consistent reconstruction of cloud-free optical imagery. The method is evaluated on the SEN12MS-CR dataset using both quantitative metrics and qualitative visual analysis under various cloud conditions.

\section{Related Work}
\label{sec:related}

Cloud removal techniques can be broadly categorized into temporal, spatial, and multi-modal approaches. Temporal interpolation methods such as STARFM \cite{Gao2006} and data fusion-based approaches \cite{Hilker2009} leverage multi-date observations to reconstruct cloud-free images. However, they are ineffective in regions with persistent cloud cover.

Spatial interpolation methods like kriging or inverse distance weighting estimate missing values using neighboring pixel statistics \cite{Roy2010}, but often blur structural boundaries in complex terrains \cite{Angel2019}.

SAR-optical fusion techniques exploit the complementary strengths of the two modalities. Traditional statistical fusion \cite{Kulkarni2020} and earlier deep learning methods like SAR-Opt-cGAN \cite{Grohnfeldt2018} and Simulation-FusionGAN \cite{Gao2020} used adversarial learning to synthesize cloud-free optical data from SAR. DSen2-CR \cite{Meraner2020} and GLF-CR \cite{Xu2022} improved accuracy using encoder-decoder structures and multi-scale supervision.

Recent advances include transformer-based architectures and uncertainty-aware learning. Notably, UnCRtainTS \cite{Ebel_10209044} combines temporal transformers with uncertainty quantification to guide cloud removal in time series data, and shows potential even in single-timestep cases. Additionally, diffusion-based \cite{rs15112861} generative models are emerging for high-fidelity satellite image generation, though their application to SAR-optical fusion is still nascent.

Despite this progress, few models explicitly differentiate cloud-occluded pixels during training. Our approach introduces a targeted, cloud-aware weighting strategy to address this gap, improving reconstruction performance in high-cloud scenarios.

\section{Proposed Framework: Cloud-Aware SAR Fusion for Enhanced Optical Image Reconstruction}
\label{sec:proposed}

This paper introduces a \textbf{Cloud-Attentive Model Framework} that integrates \textit{SAR and optical images} to reconstruct cloud-free optical imagery through a structured multi-stage process. The proposed framework incorporates \textit{attention-based feature fusion}, \textit{cloud-aware loss optimization}, and \textit{multi-resolution reconstruction} within a unified deep learning architecture to effectively address the challenges posed by cloud cover in optical satellite observations.

The process begins with a \textit{CNN-based feature extraction module}, which is applied independently to SAR and optical inputs to preserve the spectral and structural characteristics inherent to each modality. These extracted features form the basis for the subsequent fusion process.

An \textit{Attention-driven Feature Fusion mechanism} follows, combining the extracted SAR and optical features to enhance spectral-spatial integration. This module leverages dense residual encoding and transformer-based attention to facilitate robust alignment and complementary information propagation.

Upon fusion, a \textit{Reconstruction Module} is employed to synthesize a cloud-free optical image from the fused representations. This module is responsible for restoring full spatial resolution and preserving spectral fidelity in the reconstructed output.

To improve reconstruction quality in occluded areas, the framework incorporates a \textit{cloud-aware model update strategy}. This component introduces adaptive pixel-wise loss weighting, guided by a refined cloud mask, to emphasize learning on cloud-covered regions during training while ensuring generalization across cloud-free areas.

\begin{figure}[t]
	\centering
	\includegraphics[width=\linewidth]{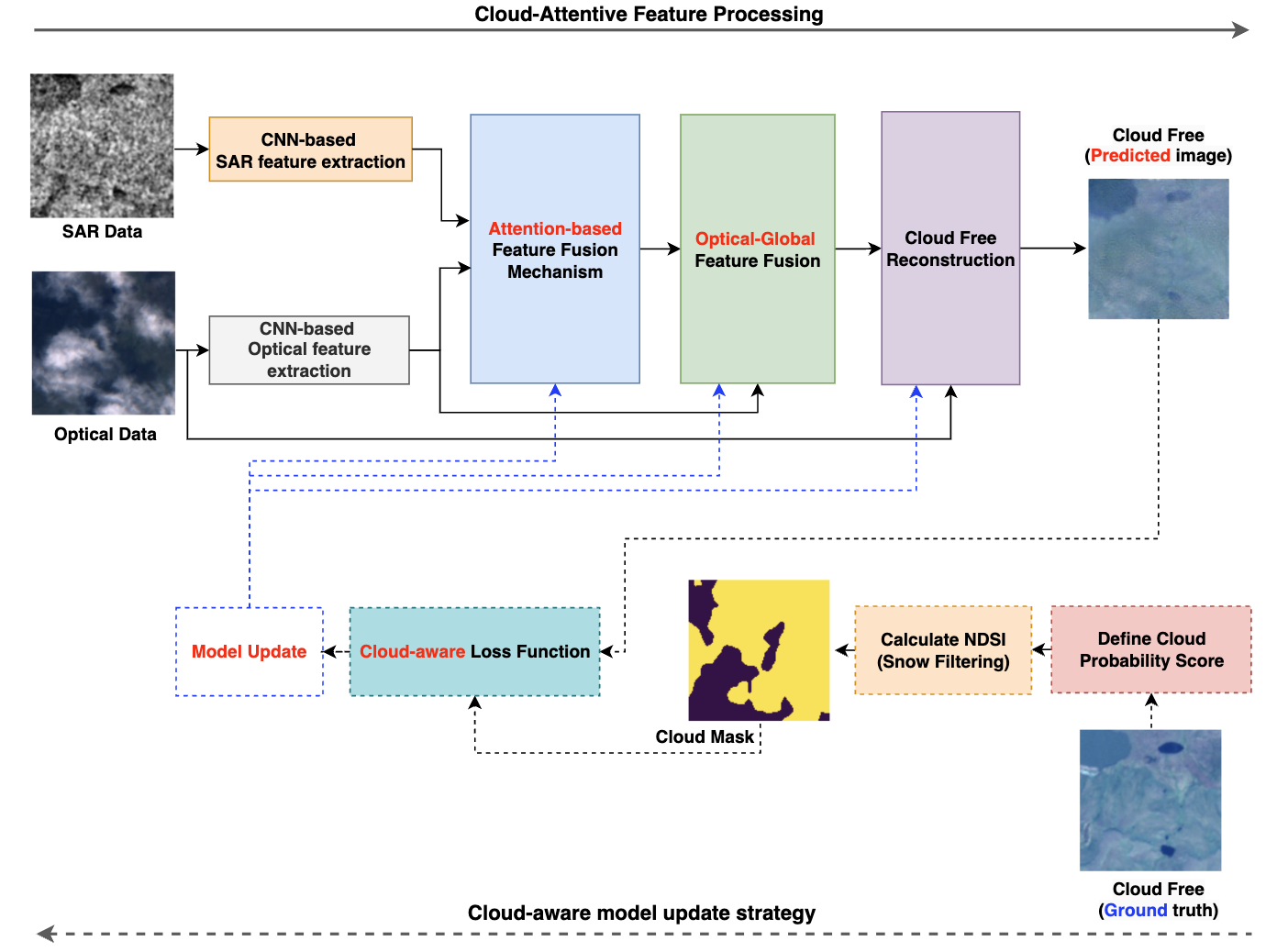}
	\caption{Workflow of the proposed \textbf{Cloud-Attentive Reconstruction Framework} for Cloud-Free Optical Imaging via SAR Fusion.}
	\label{fig:workflow}
\end{figure}

The complete processing flow—including feature extraction, multi-stage attention fusion, cloud-masked training, and spatial reconstruction—is illustrated in Figure~\ref{fig:workflow}.

\subsection{Attention-driven Feature Fusion Mechanism}
\label{subsec:Attention-driven Feature Fusion Mechanism}

\begin{figure}[t]
	\centering
	\includegraphics[width=\linewidth]{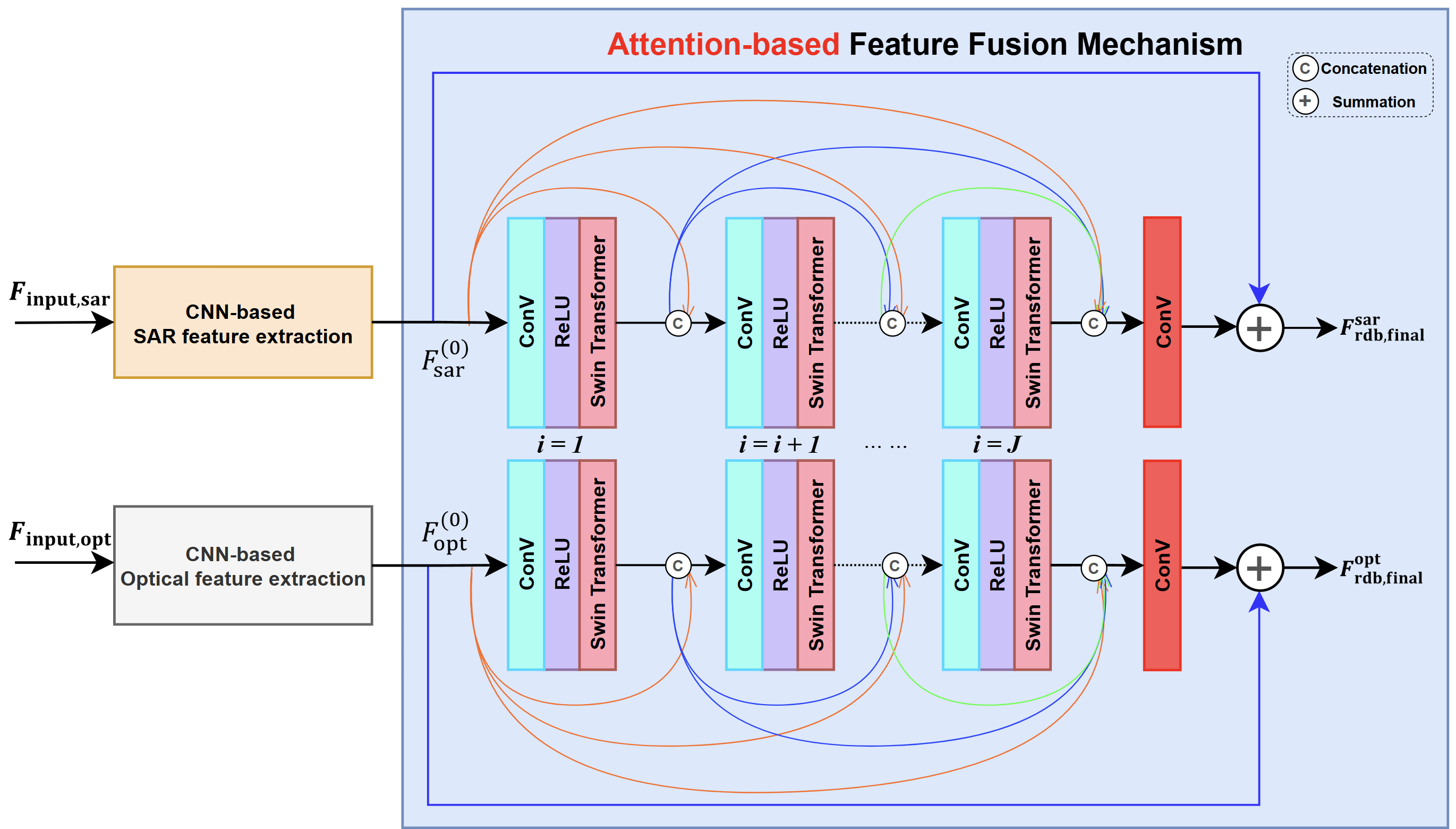}
	\caption{Proposed Attention-Based Feature Fusion Mechanism}
	\label{fig:Attention-based Feature Fusion Mechanism}
\end{figure}

This paper introduces an 	extbf{Attention-driven Feature Fusion Mechanism} to effectively integrate 	extit{Synthetic Aperture Radar (SAR)} and 	extit{optical imagery} for cloud-free optical image reconstruction. As illustrated in Figure~\ref{fig:Attention-based Feature Fusion Mechanism}, this mechanism consists of independent CNN-based feature extraction, spatial-channel reconfiguration, residual dense encoding, and transformer-based refinement.

The extraction process begins with CNN-based modules applied independently to SAR and optical inputs. This ensures preservation of their modality-specific features. SAR features offer structural insights including edges and textures, despite the presence of speckle noise \cite{ravirathinam2021neighcnn}. In contrast, optical imagery provides rich spectral information but is susceptible to cloud-induced occlusions \cite{zhou2022cloud}. Independent encoding supports retention of complementary information.

A Spatial-Channel Reconfiguration step then transforms the extracted features by redistributing spatial dimensions into channel space, enabling multiscale representation. The transformation is formulated as:

\begin{align}
	F_{\text{reconfig},m}(x,y) &= \text{Reformat}\left(F_{\text{input},m}(x,y), (C_m, H, W)\right) \\
	&\rightarrow \left(C_m' = C_m \times s^2, H' = \frac{H}{s}, W' = \frac{W}{s}\right)
	\label{Reformat}
\end{align}

where $m \in \{\text{optical}, \text{SAR}\}$, $s$ is the scaling factor, and $C_m'$ indicates the increased number of feature channels.

A Low-Level Feature Encoder (LFE) processes the reconfigured inputs through two sequential convolutional layers with ReLU activations:

\begin{align}
	F_{\text{opt}}^{(0)} &= \mathcal{F}_{\text{LFE,opt}}(F_{\text{reconfig},\text{opt}}), \\
	F_{\text{sar}}^{(0)} &= \mathcal{F}_{\text{LFE,sar}}(F_{\text{reconfig},\text{sar}})
\end{align}

Each encoder is defined as:

\begin{align}
	\mathcal{F}_{\text{LFE},m}(X) = \sigma \left( W_m^{(2)} * \sigma \left( W_m^{(1)} * X + b_m^{(1)} \right) + b_m^{(2)} \right)
\end{align}

The fusion mechanism incorporates Residual Dense Blocks (RDBs) and Swin Transformer blocks \cite{Liu2021} to enrich deep feature representations. RDBs support dense feature connectivity, while transformer blocks capture long-range dependencies and global spatial context. The Swin Transformer block is expressed as:

\begin{align}
	F_{\text{swin},j}^{(m)} = F_{\text{rdb},j}^{(m)} + \text{MLP}\left( \text{LN} \left( F_{\text{rdb},j}^{(m)} + \text{W-MSA}\left( \text{LN}(F_{\text{rdb},j}^{(m)}) \right) \right) \right)
\end{align}

Here, LN denotes Layer Normalization, W-MSA is the Window-based Multi-head Self-Attention mechanism, and MLP represents a feed-forward network with GELU activation.

After passing through $J$ RDBs and $J{-}1$ Swin Transformer blocks, the features are aggregated via a $1\times1$ convolution to compress channel dimensions:

\begin{align}
	F_{\text{rdb,final}}^{(m)} &= \text{Conv}_{1 \times 1} \Big( 
	\big[ F_{\text{swin},1}^{(m)}, \dots, F_{\text{swin},J-1}^{(m)}, F_{\text{rdb},J}^{(m)} \big] \Big), \\
	&\quad m \in \{\text{opt}, \text{sar}\}
	\label{eq:f-rdb-final}
\end{align}

This fusion mechanism enables joint exploitation of spectral and structural attributes, yielding robust and coherent features for cloud-free optical reconstruction. The architecture effectively mitigates occlusion effects and enhances the model's capability to synthesize high-resolution, spectrally faithful outputs.


\subsection{Cloud-Free Reconstruction}

\begin{figure}[h]
	\centering
	\includegraphics[width=\linewidth]{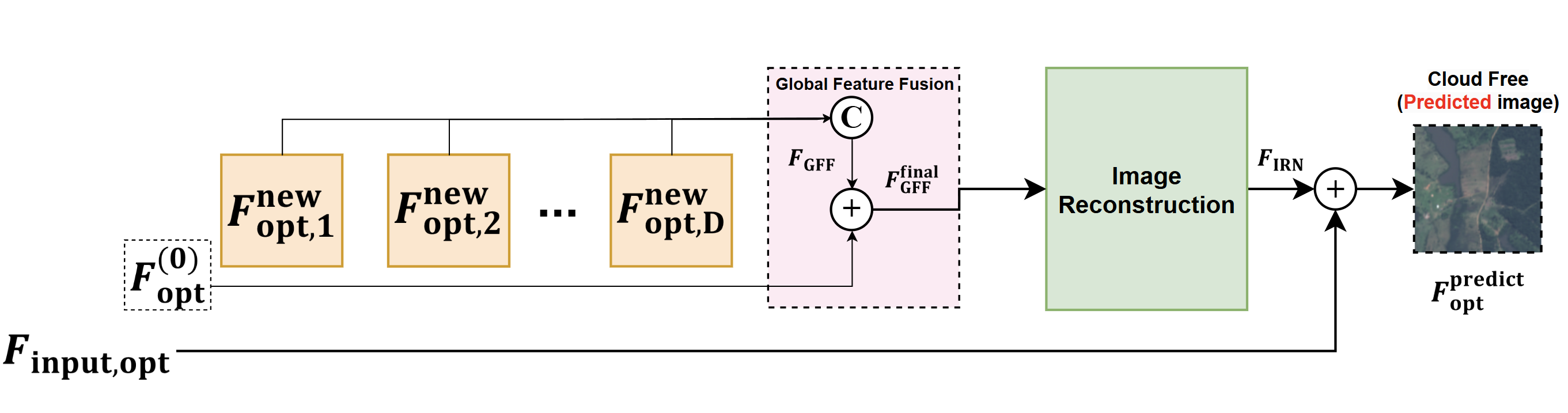}
	\caption{Cloud-Free Optical Image Reconstruction}
	\label{fig:6}
\end{figure}

The final stage of the proposed framework is responsible for synthesizing a high-fidelity cloud-free optical image by integrating features derived from SAR-optical fusion. This reconstruction phase is designed to preserve spatial structure and spectral consistency and is composed of two principal components: the Global Feature Fusion (GFF) module and the Image Reconstruction Network (IRN), as depicted in Figure~\ref{fig:6}.

To capture diverse semantic contexts, the attention-driven fusion pipeline introduced in Subsection~\ref{subsec:Attention-driven Feature Fusion Mechanism} is executed iteratively across $N$ stages. Each stage outputs refined optical feature representations, denoted as $F^{\text{new}}_{\text{opt},i}$. These intermediate features are subsequently aggregated by the GFF module to form a consolidated representation. The GFF module concatenates all $D$ intermediate feature maps and processes them using a two-step convolutional sequence:

\begin{align}
	F_{\text{GFF}} &= \text{Conv}_{3 \times 3} \left( \text{Conv}_{1 \times 1} \left( \text{Concat}\left( F^{\text{new}}_{\text{opt},1}, \dots, F^{\text{new}}_{\text{opt},D} \right) \right) \right)
\end{align}

where:
\begin{itemize}
	\item The $1\times1$ convolution reduces the dimensionality from concatenated channels.
	\item The $3\times3$ convolution enhances local continuity and refines the fused representation.
\end{itemize}

To retain the original spectral distribution, a skip connection integrates the initial optical features $F^{(0)}_{\text{opt}}$ into the output of the GFF module:

\begin{align}
	F_{\text{GFF}}^{\text{final}} = F_{\text{GFF}} + F^{(0)}_{\text{opt}}
\end{align}

Subsequently, the IRN module restores spatial resolution and transforms the high-dimensional fused features into a 13-band spectral output. This step applies inverse reformatting in conjunction with convolutional refinements:

\begin{align}
	F_{\text{IRN}} = \text{Conv}_{3\times3} \left( \text{Reformat}^{-1} \left( \text{Conv}_{3\times3}(F_{\text{GFF}}^{\text{final}}) \right) \right)
\end{align}

where:
\begin{itemize}
	\item $\text{Reformat}^{-1}$ reverts spatial-channel compression, restoring the original resolution.
	\item Convolutional filters sharpen the reconstruction and suppress artifacts.
\end{itemize}

To further refine the reconstruction and maintain consistency with the original input, a residual connection adds the cloudy optical image:

\begin{align}
	F_{\text{opt}}^{\text{predict}} = F_{\text{IRN}} + F_{\text{input,opt}}
\end{align}

This residual learning strategy ensures that fine spectral details are preserved while improving predictions in cloud-covered regions. The synergy between GFF and IRN completes the cloud removal pipeline by generating spatially coherent and spectrally faithful optical reconstructions.


\subsection{Cloud-Aware Model Update Strategy}

A key innovation of the proposed framework lies in the \textit{cloud-aware model update strategy}, which addresses the challenge of reconstructing cloud-covered optical regions. This strategy incorporates an enhanced cloud detection procedure and an adaptive loss weighting mechanism to prioritize learning in occluded areas while maintaining consistency in cloud-free regions.

The cloud detection module adopts an enhanced Sen2Cor-based approach \cite{Main2017}, leveraging the multispectral characteristics of Sentinel-2 imagery. Since clouds exhibit strong reflectance in visible and near-infrared (NIR) bands—including coastal blue ($B_1$), blue ($B_2$), green ($B_3$), and cirrus ($B_{10}$)—a cloud probability score $S_c(i,j)$ is defined as follows:

\begin{align}
	S_c(i,j) = \min\Bigg(&
	\frac{B_2(i,j)-0.1}{0.5-0.1}, \notag \\
	&\frac{B_1(i,j)-0.1}{0.3-0.1}, \notag \\
	&\frac{B_{10}(i,j)+B_1(i,j)-0.15}{0.2-0.15}, \notag \\
	&\frac{B_4(i,j)+B_3(i,j)+B_2(i,j)-0.2}{0.8-0.2}
	\Bigg)
\end{align}

A binary mask $M(i,j)$ is derived using a threshold $T_{\text{cloud}} = 0.2$:
\begin{align}
	M(i, j) = \begin{cases} 
		1, & \text{if } S_c(i, j) > T_{\text{cloud}} \\
		0, & \text{otherwise}
	\end{cases}
\end{align}

To eliminate false positives caused by snow, the Normalized Difference Snow Index (NDSI) is computed:
\begin{align}
	NDSI(i,j) &= \frac{B_3(i,j) - B_{11}(i,j)}{B_3(i,j) + B_{11}(i,j)} \\
	M'(i, j) &= M(i,j) \times \mathbb{1}[NDSI(i,j) \leq 0.6]
\end{align}

\begin{figure}[h]
	\centering
	\includegraphics[width=0.8\linewidth]{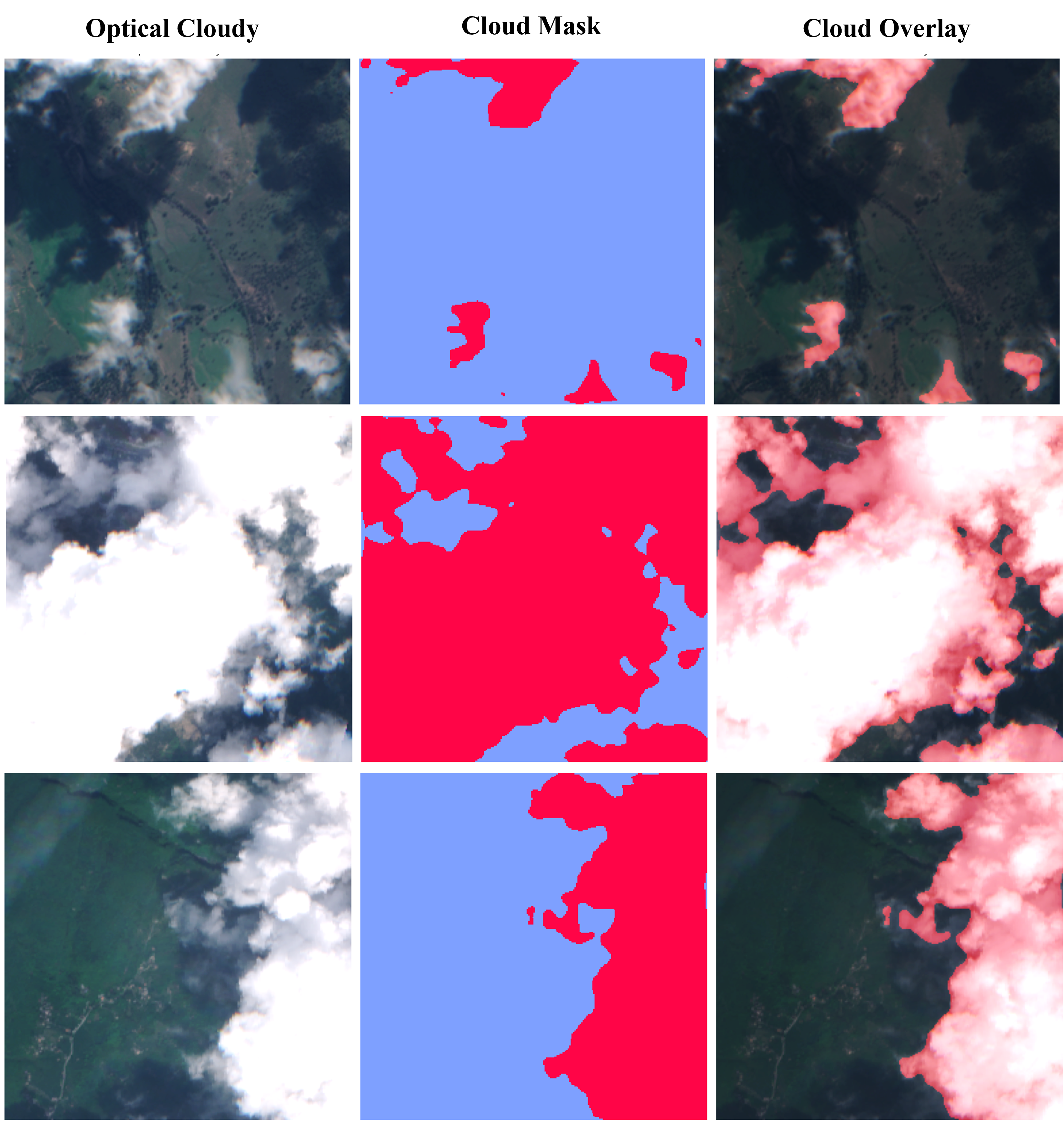}
	\caption{Results of cloud masking algorithm $M'(i,j)$}
	\label{fig:cloud-mask}
\end{figure}

Figure~\ref{fig:cloud-mask} presents a visual result of the refined cloud mask $M'(i,j)$.

To dynamically prioritize occluded regions during training, an adaptive weight function is introduced:
\begin{align}
	W_{\text{cloud}}(x,y) = \alpha \cdot M'(x,y) + (1-\alpha) \cdot (1-M'(x,y))
\end{align}

where $\alpha = 0.8$ emphasizes cloudy pixels by assigning higher weights.

The final cloud-aware loss function combines Mean Squared Error (MSE) and Structural Similarity Index Measure (SSIM):
\begin{align}
	\mathcal{L}_{\text{final}}(x,y) = W_{\text{cloud}}(x,y) \cdot \left( \lambda_1 \mathcal{L}_{\text{MSE}}(x,y) + \lambda_2 \mathcal{L}_{\text{SSIM}}(x,y) \right)
\end{align}

where $\lambda_1 = \lambda_2 = 0.5$ ensures balanced emphasis between pixel accuracy and perceptual structure.

This adaptive loss formulation promotes accelerated convergence and improved fidelity in heavily clouded regions, while ensuring the preservation of spatial and spectral consistency across the entire reconstructed image.

\section{Experiment Results}
\label{sec:results}

To validate the effectiveness of the proposed framework, experiments are conducted on the SEN12MS-CR dataset \cite{Ebel2024}, which provides co-registered Sentinel-1 SAR (VV, VH) and Sentinel-2 multi-spectral (13-band) imagery. The dataset includes 169 globally distributed locations with multi-seasonal coverage, capturing diverse cloud conditions and surface reflectance properties. Each image patch has a spatial resolution of 10m and a size of $256 \times 256$ pixels. The dataset features a broad cloud coverage distribution ranging from light to heavily occluded scenes (10–90\%+), ensuring comprehensive evaluation across challenging scenarios.

The dataset is split into $80\%$ training, $10\%$ validation, and $10\%$ testing sets, with no spatial overlap between subsets. All results are averaged over three independent training runs using different random seeds to account for model variance and ensure statistical reliability.

The implementation is based on PyTorch 2.0 and optimized using the Adam optimizer with an initial learning rate of $1 \times 10^{-5}$. Performance is assessed using three widely accepted image reconstruction metrics: Peak Signal-to-Noise Ratio (PSNR), Structural Similarity Index Measure (SSIM), and Mean Absolute Error (MAE). Definitions are as follows:

\begin{align}
	\text{PSNR}(O_\text{gt}, \hat{O}) &= 10 \log_{10} \left( \frac{255^2}{\text{MSE}} \right) \\
	\text{MAE} &= \frac{1}{N} \sum_{i=1}^N \left| \hat{O}(x_i, y_i) - O_\text{gt}(x_i, y_i) \right|
\end{align}

The proposed framework is compared against five baselines: SAR-Opt-cGAN \cite{Grohnfeldt2018}, Simulation-FusionGAN \cite{Gao2020}, DSen2-CR \cite{Meraner2020}, GLF-CR \cite{Xu2022}, and the recent UnCRtainTS \cite{Ebel2023}. Results are presented in Table~\ref{tab:comparison}.

\begin{table}[h]
	\centering
	\footnotesize
	\begin{tabular}{lccc}
		\toprule
		\textbf{Method} & \textbf{PSNR (dB)}$\uparrow$ & \textbf{SSIM}$\uparrow$ & \textbf{MAE}$\downarrow$ \\
		\midrule
		SAR-Opt-cGAN \cite{Grohnfeldt2018} & 25.29 & 0.764 & 0.043 \\
		Simulation-FusionGAN \cite{Gao2020} & 24.55 & 0.701 & 0.046 \\
		DSen2-CR \cite{Meraner2020} & 27.38 & 0.874 & 0.032 \\
		GLF-CR \cite{Xu2022} & 29.73 & 0.885 & 0.025 \\
		UnCRtainTS \cite{Ebel2023} & 30.15 & 0.880 & 0.023 \\
		\textbf{Proposed Method} & \textbf{31.01} & \textbf{0.918} & \textbf{0.017} \\
		\bottomrule
	\end{tabular}
	\caption{Comparison of cloud removal methods using PSNR, SSIM, and MAE.}
	\label{tab:comparison}
\end{table}

As shown, the proposed method achieves the highest PSNR, SSIM, and the lowest MAE among all baselines. Compared to the most recent and competitive model UnCRtainTS, the proposed framework achieves a PSNR improvement of $+0.86\,\text{dB}$ and reduces MAE by over $26\%$. These results demonstrate the effectiveness of the attention-driven fusion and cloud-aware learning strategy.

To isolate the contribution of the proposed cloud-weighted loss function, an ablation study was conducted by disabling the weighting mechanism. Results showed a drop in PSNR from $31.01$ to $29.58\,\text{dB}$ and an increase in MAE from $0.017$ to $0.021$, highlighting the importance of adaptive emphasis on cloud-occluded regions during training. SSIM also declined from $0.918$ to $0.891$, indicating reduced structural fidelity.

\begin{figure}[h]
	\centering
	\includegraphics[width=\linewidth]{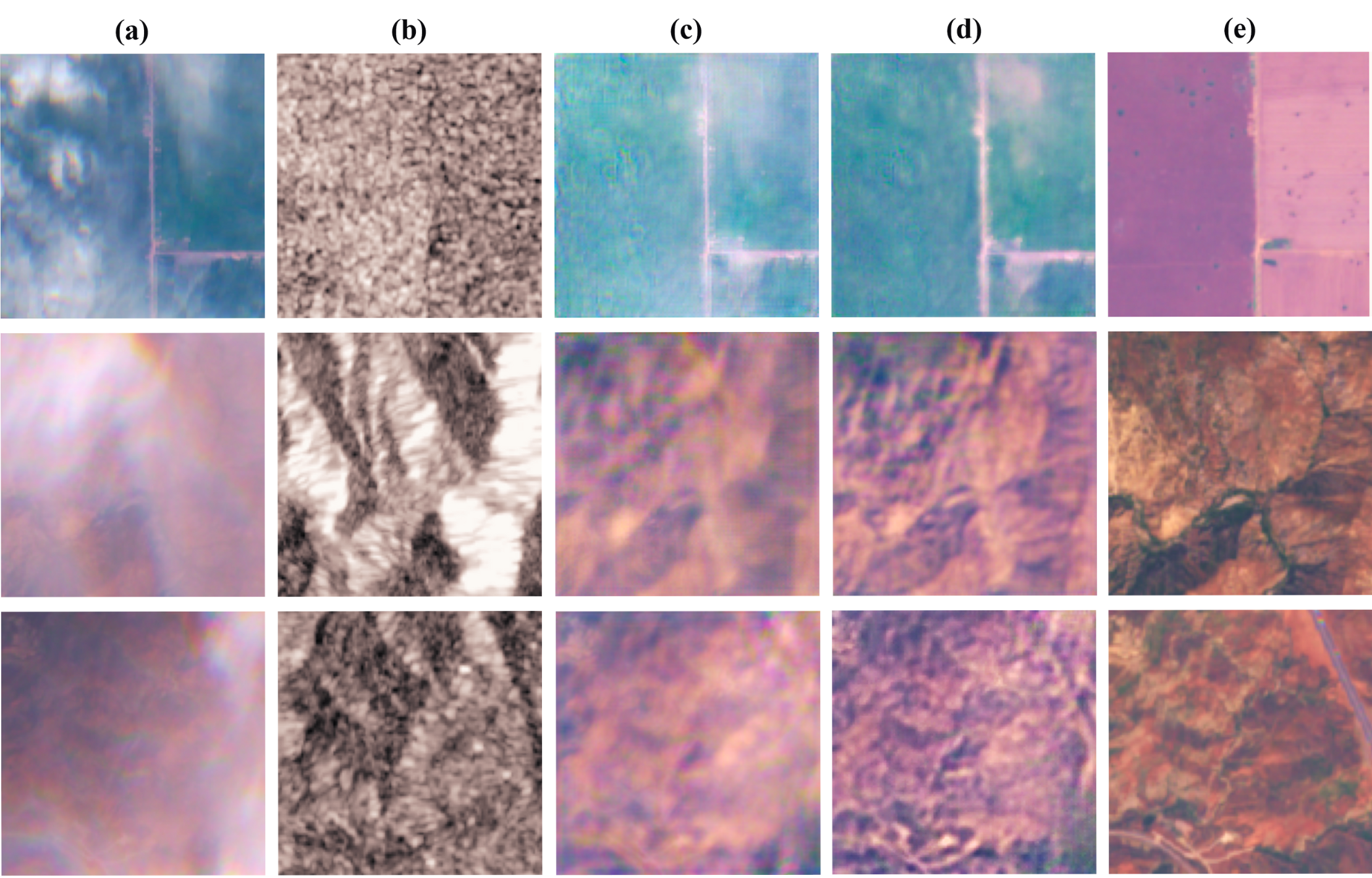}
	\caption{Visual comparison for $\sim50\%$ cloud coverage. (a) Cloudy input, (b) SAR, (c) GLF-CR \cite{Xu2022}, (d) Proposed method, (e) Ground truth.}
	\label{fig:results1}
\end{figure}

\begin{figure}[h]
	\centering
	\includegraphics[width=\linewidth]{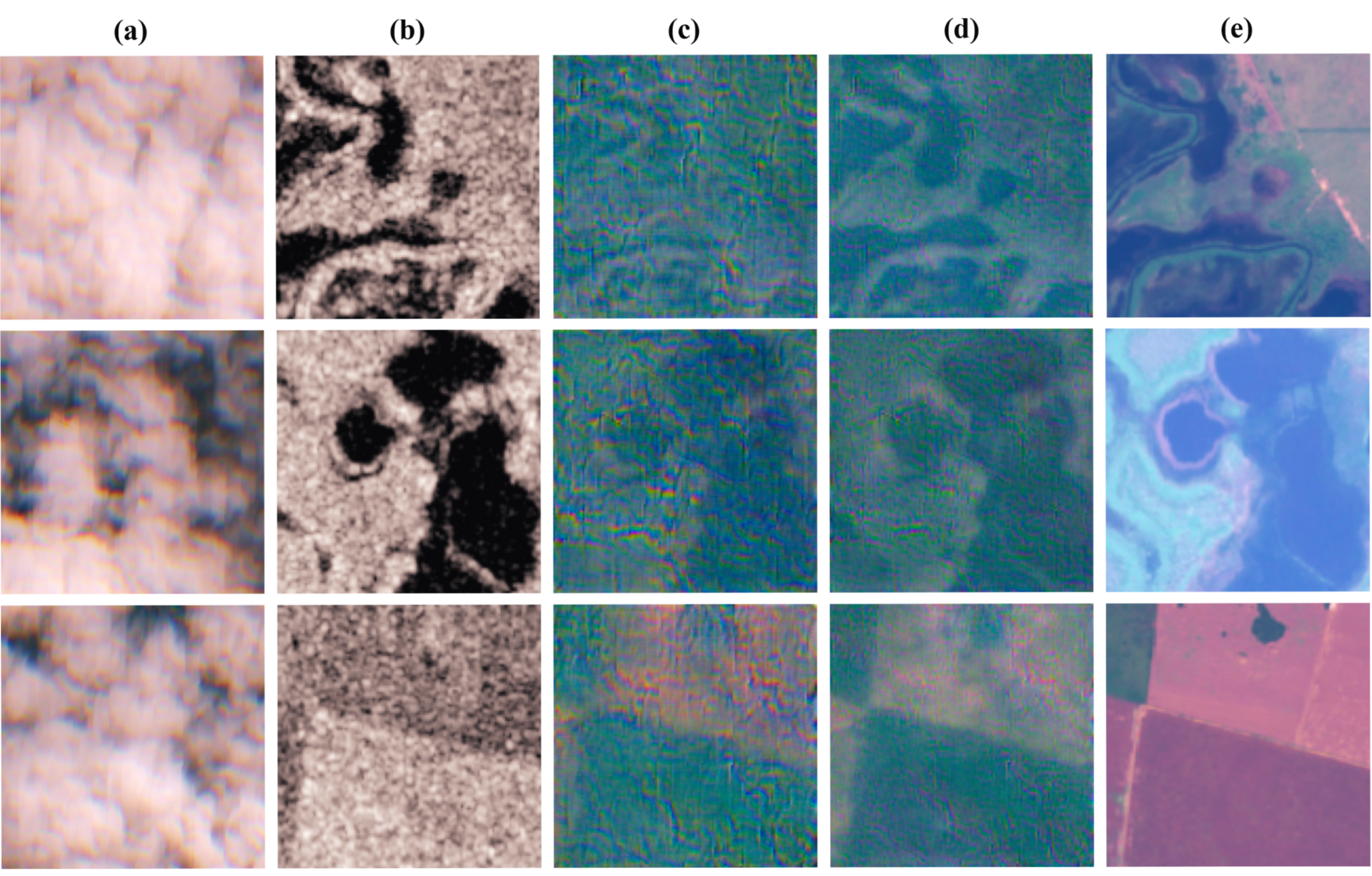}
	\caption{Visual comparison for $>90\%$ cloud coverage. (a) Cloudy input, (b) SAR, (c) GLF-CR \cite{Xu2022}, (d) Proposed method, (e) Ground truth.}
	\label{fig:results2}
\end{figure}

Qualitative comparisons in Figures~\ref{fig:results1} and~\ref{fig:results2} illustrate the superiority of the proposed method over GLF-CR across both moderate ($\sim50\%$ cloud) and highly occluded ($>90\%$ cloud) scenarios. The proposed method restores fine details and spectral consistency more faithfully, even under severe cloud interference.

\section{Conclusion and Discussion}
\label{sec:conclusion}

This paper presents the \textit{Cloud-Attentive Reconstruction Framework}, a unified model for generating cloud-free optical imagery by fusing SAR and optical data using attention-guided deep learning. The proposed method introduces a multi-stage attention-driven feature fusion mechanism that effectively integrates complementary spectral and structural cues. Additionally, a cloud-aware model update strategy dynamically emphasizes cloud-occluded regions during training, enhancing reconstruction quality. Experimental results on the SEN12MS-CR dataset confirm that the method consistently outperforms state-of-the-art baselines across PSNR, SSIM, and MAE metrics, while visual comparisons highlight superior restoration of fine details under various cloud conditions.

Despite the strong performance, several limitations remain. The reliance on SAR data introduces sensitivity to speckle noise and geometric distortions, particularly in heterogeneous or mountainous terrains. While the SEN12MS-CR dataset provides well-aligned SAR and optical imagery, such co-registration may not be guaranteed in practical deployments. In real-world scenarios, differences in spatial resolution, acquisition geometry, or sensor timing can lead to misalignment, which can hinder effective fusion. Addressing this requires incorporating preprocessing steps such as feature-based registration, mutual information alignment, or deep learning-based co-registration into the operational pipeline.


In summary, the proposed framework offers a scalable and effective solution for optical cloud removal, with promising applications in environmental monitoring, agricultural assessment, and disaster response. Continued work toward generalization, alignment robustness, and computational efficiency will further enhance its practicality for spaceborne Earth observation missions.

{
	\small
	\bibliographystyle{IEEEtran}
	\bibliography{main}  
}
\end{document}